\documentclass[pdflatex,sn-mathphys]{sn-jnl}
\usepackage{subfig}
\usepackage[capitalise]{cleveref}
\usepackage{hyperref}
\usepackage{multirow}
\usepackage{tabularx}
\usepackage{interval}
\usepackage{placeins}
\usepackage{listings}
\usepackage{booktabs}

\jyear{2022}%

\theoremstyle{thmstyleone}%
\theoremstyle{thmstyletwo}%

\theoremstyle{thmstylethree}%

\begin{document}


\title[ARCOR2: End-User Management of Robotic Workplaces using AR]{ARCOR2: Framework for Collaborative End-User Management of Industrial Robotic Workplaces using Augmented Reality}


\author*[1]{\fnm{Michal} \sur{Kapinus}}\email{ikapinus@fit.vut.cz}

\author[1]{\fnm{Zdeněk} \sur{Materna}}\email{imaterna@fit.vut.cz}

\author[1]{\fnm{Daniel} \sur{Bambušek}}\email{bambusekd@fit.vut.cz}

\author[1]{\fnm{Vítězslav} \sur{Beran}}\email{beranv@fit.vut.cz}

\author[1]{\fnm{Pavel} \sur{Smrž}}\email{smrz@fit.vut.cz}

\affil[1]{\orgdiv{Faculty of Information Technology}, \orgname{Brno University of Technology}, \orgaddress{\street{Božetěchova 1/2}, \city{Brno}, \postcode{612 00}, \country{Czech Republic}}}

\abstract{This paper presents a novel framework enabling end-users to perform the management of complex robotic workplaces using a tablet and augmented reality. The framework allows users to commission the workplace comprising different types of robots, machines, or services irrespective of the vendor, set task-important points in space, specify program steps, generate a code, and control its execution. More users can collaborate simultaneously, for instance, within a large-scale workplace. Spatially registered visualization and programming enable a fast and easy understanding of workplace processes, while high precision is achieved by combining kinesthetic teaching with specific graphical tools for relative manipulation of poses. A visually defined program is for execution translated into Python representation, allowing efficient involvement of experts. The system was designed and developed in cooperation with a system integrator based on an offline printed circuit board testing use case, and its user interface was evaluated multiple times during the development. The latest evaluation was performed by three experts and indicates the high potential of the solution.}

\keywords{visual programming, augmented reality, collaborative robot, end-user programming}

\pacs[Category]{6}



\maketitle

\section{Introduction}


Most often, robots are used for highly repetitive tasks. For instance, the automotive industry's production line is programmed once and works for several years without major changes. On the other hand, in Small and Medium Enterprises (SMEs), production changes more often. Each batch can be customized, and the robot may be used for multiple purposes. If there is a need to reprogram the robot, the company needs its own highly skilled employee or must use the services of external suppliers, which might be expensive or not flexible enough. Moreover, end-users might not be experts in programming or robotics but certainly might have invaluable task-domain knowledge. Therefore, there is a trend towards allowing end-users to program the robots: robot manufacturers are introducing simplified teach pendants and collaborative robots able to be hand-taught, third parties are developing visual programming tools, etc. Still, existing solutions have several pitfalls:

\begin{itemize}
    \item Use proprietary language and development environment.
    \item Require textual coding or are not expressive enough.
    \item Do not allow to work within the task space.
    \item Missing visualization of spatial information.
    \item Limited to programming robots (typically just one).
\end{itemize}

To overcome existing limitations, we present the ARCOR2 framework\footnote{Source code and documentation (including integration for ABB YuMi, Dobot M1 and Dobot Magician robots and for Kinect Azure sensor) are available at \href{https://github.com/robofit/arcor2}{github.com/robofit/arcor2}.} which enables end-users to perform complete management of a robotic workplace or a production line: initial setup, programming, adaptation, releasing to production, controlling execution, etc. Its user interface can be seen as a universal teach pendant for all robot types, machines, or APIs where a new device or service can be integrated by writing a custom plugin in Python. This integrative approach eliminates the need to undergo training for the interface of each device involved. The user interface is designed for commodity tablets. It utilizes augmented reality (AR) to visualize program data, including spatial points, individual program instructions, and even a logical flow of the program. One tablet can be used to manage multiple workplaces.
\begin{figure}
  \includegraphics[width=0.88\textwidth]{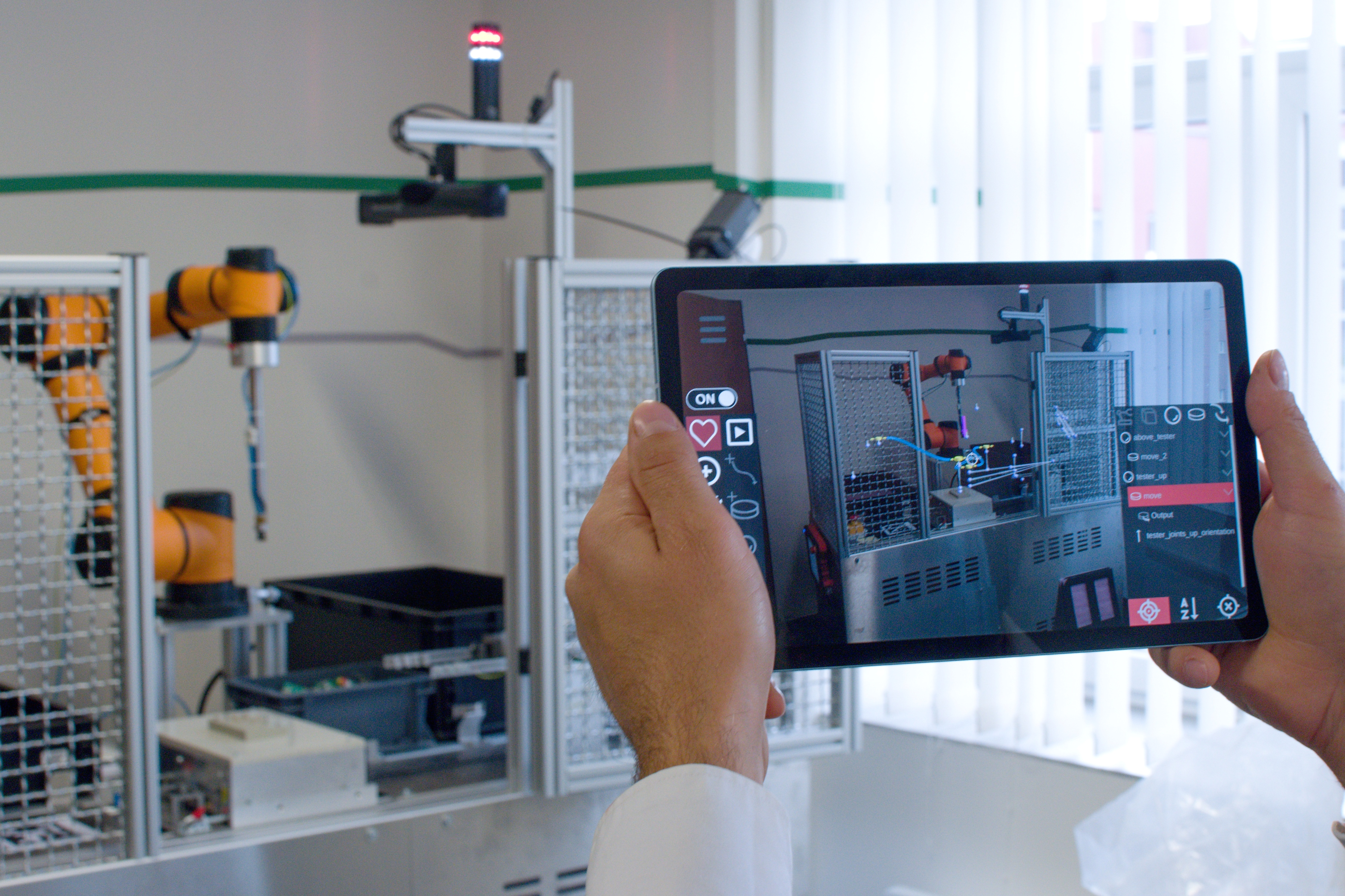}
  \centering
  \caption{A user observes the program for the PCB testing through the tablet.}
  \label{fig:intro-aubo}
\end{figure}

The framework was developed in cooperation with a system integrator\footnote{\href{https://www.kinali.cz/en/}{www.kinali.cz/en}}, who provided the PCB testing use case and corresponding testing site, participated in gathering the requirements and continuous testing of both the backend part and user interface of ARCOR2. The company has also developed an integration for its services and robots.

This paper is both culmination of the three-year project as well as our long-term research in the field of human-robot interaction within industrial environments, which research questions could be summarized as follows:

\begin{itemize}
    \item \textbf{Q1} -- How can interaction be brought from a 2D screen back to the 3D world?
    \item \textbf{Q2} -- Is (affordable) handheld AR a viable modality?
    \item \textbf{Q3} -- How to allow precise work with imprecisely registered AR devices?
    \item \textbf{Q4} -- What is the proper level of abstraction for end-users?
\end{itemize}

The goals of the paper are to summarize our work within the field and on the framework, to present the framework's general applicability and utility, perform an overall evaluation (as previously published evaluations were focused on specific aspects), and share the gained experience with the community\footnote{For a video summary of this paper, please see \href{https://youtu.be/RI1uiIEiPK8}{youtu.be/RI1uiIEiPK8}.}.

\label{sec:related-work}
\section{Related work}




In industrial settings, an ordinary worker operates the robot most often at Level 0 (bystander) as defined in \cite{ionescu2019participatory}, while the higher levels are handled by a specially trained person or an external expert. With the increasing spread of collaborative robots, rising needs for flexible or customizable production, and deployment of robots into smaller industries, there is a trend toward allowing end-users to program robots.

\subsection{Established Solutions}

Although some off-the-shelf teach pendants offer a certain form of simplified programming, the usability seems to be rather low \cite{schmidbauer2020teaching} due to  missing visualization, inability to use common syntax structures as conditions and loops, high mental and physical demands or lack of tools for debugging \cite{ajaykumar2020user, weiss2016first, huang2020contextual, ionescu2019participatory}. 

Moreover, pendants are vendor-specific and limited to programming robots. On the other hand, offline programming tools such as RoboDK\footnote{\href{https://robodk.com/}{robodk.com}} or ABB RobotStudio\footnote{\href{https://new.abb.com/products/robotics/en/robotstudio}{new.abb.com/products/robotics/en/robotstudio}} offer comprehensive functionality but require extensive training. Additionally, PC-based and pendant-like user interfaces are, in general, likely not optimal for end-user programming, as they imply continuous switching of a user's attention \cite{weiss2016first} and therefore induce high cognitive and attention-related workload. Kinesthetic teaching is often employed to allow users to set target poses or waypoints intuitively and, therefore, simplify programming. However, depending on the stiffness and size of a particular robot, it could be physically demanding and not desirable by users~\cite{huang2020contextual, ajaykumar2020user}.




\subsection{Proposed Approaches}

Many alternative methods for simplified programming and even third-party complete solutions were proposed. Some of them are not intended as comprehensive tools but rather focused on a specific task or aspect of the process or limited to a certain robot. The simplification is usually achieved through some form of visual programming~\cite{paxton2017costar, stenmark2017simplified, gao2019pati, mayr2021considerations, huang2017code3}, spatial visualization enabling the user to work within the task context~\cite{gao2019pati, yigitbas2021simplifying, thoo2021online}, commonly combined with a kinesthetic teaching~\cite{paxton2017costar, liang2021iropro, rodamilans2016comparison} and/or perception \cite{huang2017code3}. Unfortunately, there still exist many limitations. Only a small portion of evaluations are carried out on non-trivial use-cases as in \cite{mayr2021considerations}, or contain comparison with an existing method as in \cite{perzylo2016intuitive, stenmark2017simplified, gao2019pati}. Often, there is only a simplified method available, which precludes the possibility of (remote) expert intervention, where it can be assumed that the expert prefers to work with source code. The issue could be, for instance, solved by generating the source code from visual representation \cite{ladeira2021robmex, yigitbas2021simplifying} and probably optimally by bidirectional synchronization between those two.

\subsection{Visualization Methods}

AR seems to be a promising visualization method for simplified programming on a high level of abstraction. It may enable users to work within the task space and avoid superfluous attention switches, mental transformations, and the  related workload. However, only a few approaches allow programming in AR \cite{gadre2019end, yigitbas2021simplifying, gao2019pati} or provide the ability to set up a workspace \cite{thoo2021online} and therefore do not require an additional intermediate user interface.

Often, AR is used only as an extension, e.g., to visualize robot trajectories, and in general, there is a lack of tools for precise manipulation of robot or virtual elements, which is necessary for industrial use cases. The AR may be, for instance, realized by spatial projection \cite{gao2019pati, materna2018interactive}, which is indeed limited to visualization on surfaces. Head-up stereoscopic displays can convey depth but, on the other hand, are expensive, offer a limited field of view, and require learning of unconventional control (e.g., gestures). Usage of hand-held devices leads to problems with missing depth perception \cite{thoo2021online}; however, those are affordable, portable, and easy to use.

\subsection{Summary}

As seen from the previous related work overview, many approaches exist that aim to lower the barrier to entry the robot programming by various means. However, there is a lack of comprehensive yet simple environments that allows end-users to perform all tasks and steps necessary for industrial-like applications. Also, it has to be considered that modern workplaces may contain not only a robot but multiple (programmable) machines or special-purpose devices. With Industry 4.0, there will also be a rising need to communicate with various services through their APIs.

\section{Use Case}
\label{sec:use-case}

Although the framework was designed to be general, the initial motivation came up from the use case of offline product testing in an SME company, where relatively small batches are tested, and items' storage is highly variable. Therefore a program has to be adapted approximately once a week. The process consists of picking up either unorganized items from crates or organized ones from blisters, applying printed barcodes, putting items into a tester, starting the test, and placing items in boxes according to a functional test result. So far, the process was done by human operators, but their time was not used efficiently, as they were idle for up to several minutes while the test ran. Moreover, the work is highly stereotypical. The goal was to optimize the use of a qualified workforce as most of the operators, after robotizing the process, could be reassigned to more creative work. The rest could be trained to adapt programs of testing workplaces when needed and supervise multiple workplaces during execution. 

\begin{figure}
  \includegraphics[width=0.68\textwidth]{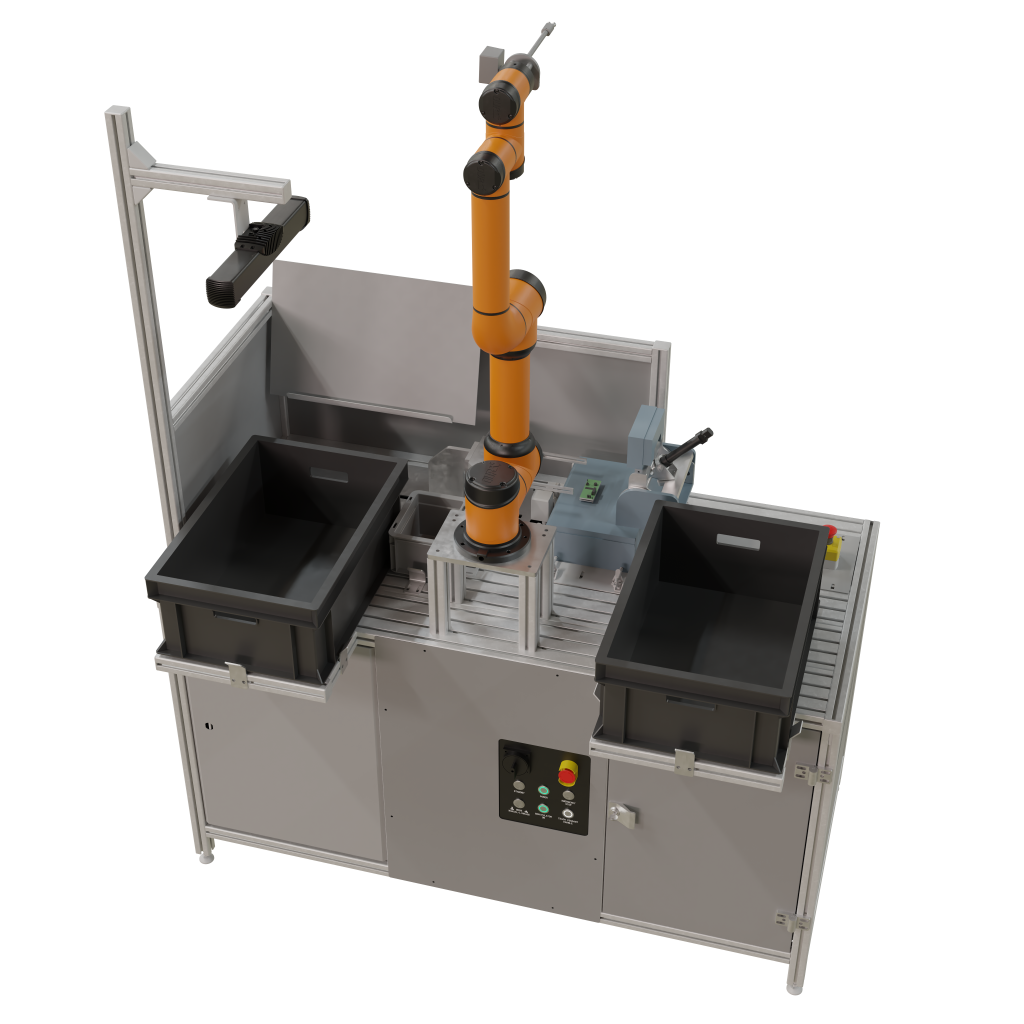}
  \centering
  \caption{Render of a PCB testing workplace with the Ensenso 3D camera for bin-picking, 6~DoF Aubo i5 robot, 2~DoF custom-build robot, functional tester, barcode reader, and printer, source, and target boxes.}
  \label{fig:tio-render}
\end{figure}

A robotic workplace designed for this use case (for its visualization, see \cref{fig:tio-render}) consists of one standard collaborative robot, one purpose-built robot for fine manipulation with PCBs, a 3D camera, a functional tester, and several other physical devices. Moreover, it must communicate with several software systems such as ERP and SCADA. In the case of the traditional approach, the collaborative robot would be programmed using a teach pendant, and a PLC would operate the custom one and all the other devices. To adapt such a heterogeneous workplace to a new product, a highly trained operator would be needed. In the proposed approach, the system integrator will develop an integration for all devices into the ARCOR2 system, providing functionality on the optimal level of abstraction for the task. The system integrator will also do the initial setup of each workplace. Then, changes can be done by a trained operator or remotely by an expert programmer. The main advantage for the end-user is just one configuration, programming, visualization, and control interface.

Based on a comprehensive discussion with the project partner, a set of requirements for the system were defined:

\begin{enumerate}
    \item Convenient integration of new robots, machines, and services with variable levels of abstraction. 
    \item Support collaboration between end-users and experts.
    \item Ability to manage (perform CRUD operations on):
    \begin{enumerate}
        \item Setups of the workplaces (available objects, their locations, and parameters).
        \item Important points in space and associated data.
        \item Program steps and their parameters.  
        \item Self-contained executable snapshots of programs.
    \end{enumerate}
    \item Robot as a source of precision.  
    \item Control and visualization of an execution state.
    \item Debugging capabilities.
\end{enumerate}

We have also defined different user roles divided into two categories (see \cref{tab:expected-users}). For each role, there are different responsibilities and needs. 


%

\begin{table}[]
\centering
\caption{Expected types of users, divided into two main categories.}
\label{tab:expected-users}
\begin{tabular}{llp{0.25\linewidth}p{0.25\linewidth}} 
\toprule
\textbf{Category} & \textbf{Role} & \textbf{Responsibilities} & \textbf{Principal needs} \\ 
\cmidrule(lr){1-4}
\multirow{3}{*}{End User} & Operator & Manages program execution, solves simple problems. & Visualization of execution state, controls to start/stop the program, notifications on errors. \\ \cmidrule(lr){2-4}
                        & Standard User & Able to create a simple program or adapt an existing one.       & Program management (edit, copy, etc.), tools to edit spatial points and program steps.               \\ \cmidrule(lr){2-4}
                        & Advanced User  & Able to create complex programs visually, can write simple code. & Visual definition of advanced concepts, simple and well-documented programming API.\\ \cmidrule(lr){1-4}
\multirow{2}{*}{Expert} & Technician & Performs initial setup, called when a serious problem occurs. & Debugging tools, entering exact numbers. \\ \cmidrule(lr){2-4}
                        & Programmer & Integrates new devices, creates new functionality. & Well-defined processes, generality, and reusability of code. \\ 
\bottomrule
\end{tabular}
\end{table}

\label{sec:system-design}
\section{System Design}

The defined requirements mainly give the design of the framework. However, it was also influenced by the knowledge gained during the development and evaluations of its previous generation called ARCOR \cite{materna2018interactive}, which utilized spatial AR and a touch-enabled table for user interaction and was focused mainly on table-top scenarios. Later, to overcome the inability to visualize information in free 3D space, HoloLens were integrated \cite{bambusek2019combining}. Although ARCOR was successfully evaluated in an industrial use case~\cite{kapinus2020end}, its limitations (mainly complicated integration of new devices and program instructions) lead us to the development of the next generation.

\subsection{Terminology}

\textbf{Object Type} -- plugin into the system that represents and provides integration with a particular type of real-world object, e.g., a certain type of robot or a virtual object such as cloud API. It is written in Python and can benefit from (multiple) inheritance to extend or share functionality. A set of built-in base classes is available, representing, e.g., a generic robot or a camera and its required API. It could be associated with a model (representing both collision and visual geometry), which might be a primitive or a mesh.

\textbf{Action Object} -- an instance of an object type within the workplace, defined by its unique ID (UID), human-readable name, pose (optionally), and parameters (e.g., API URI, serial port, etc.).

\textbf{Scene} -- set of action objects, represents a workplace, its objects, and spatial relations. 

\textbf{Action Point} -- a spatially anchored container for orientations, robot joint configurations, and actions. The container's position, together with orientation, comprises a pose usable, e.g., as a parameter for robot action.

\textbf{Action} -- method of an object type exposed to the AR environment. A named and parameterized action is called an action instance. Actions may be implemented on different levels of complexity according to the application needs and the target end-users competencies. However, to lower program complexity and reduce training time, the actions should be preferably high-level and provide configurable skill-like functionality. Such actions can be seen as equivalent to reusable skills used, e.g., in \cite{stenmark2017simplified, liang2021iropro}.

\textbf{Project} -- set of action points, may contain logic definition. The project is associated with a scene.

\textbf{Execution Package} -- a self-contained executable snapshot of a project, created when there is a need to test the whole task or release a project into a production environment. The fact that the package is self-contained allows users to make further changes in the scene or project without any influence on existing packages.

\textbf{Main Script} -- contains a logic of the project, which may be defined visually or written manually with the help of a set of generated classes providing access to project data as, e.g., defined action points.

\subsection{Integrating New Devices}


A new device is integrated into the system by implementing an object type (Python wrapper) that is based on some of the provided abstract base classes and is dynamically loaded into the system.

For instance, there is an abstract \texttt{Robot} class, and all object types representing particular robots are derived from it. It has a set of basic abstract methods representing mandatory, or robot's minimal functionality (e.g., a method to get the end effector pose), that must be implemented. Then, there is a set of methods that may or may not be implemented based on the available functionality of the particular robot (e.g., a method for forward and inverse kinematics or for toggling the hand teaching mode). After the wrapper is loaded, the system performs static analysis to determine in advance which optional methods are available, and based on that, certain functionality is or is not made available to the user.

There are two main possibilities of how an object type could be interfaced with a real-world object, e.g., a robot:

\begin{enumerate}
    \item Directly -- if the robot provides API with the necessary functionality, the object type may communicate with it directly.
    \item Through an intermediate service -- for instance, if the robot lacks motion planning capability, there might be a ROS-based container between the robot and the object type.
\end{enumerate}

In both cases, the object type is the main provider of all actions available to the user, regardless of the interface between the object type and the real-world object.

\label{subsec:architecture}
\subsection{Architecture}  

\begin{figure}
  \includegraphics[width=0.98\textwidth]{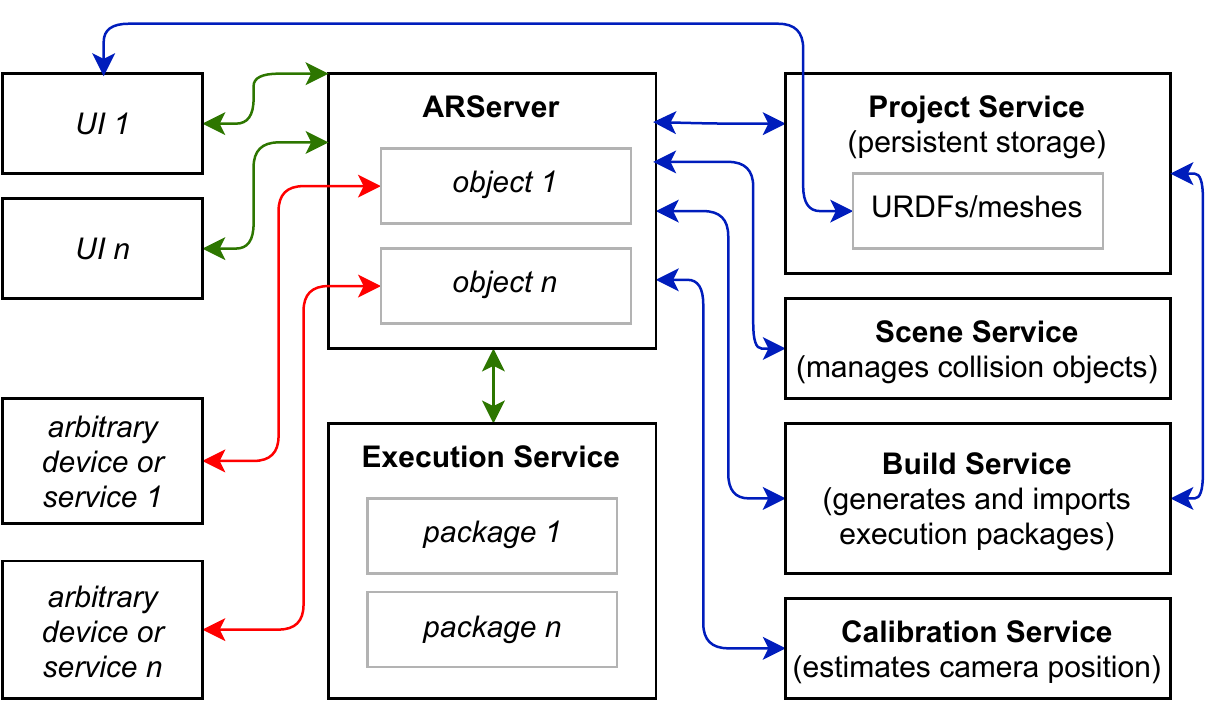}
  \centering
  \caption{Block diagram of the system in a state when object instances are created in the ARServer (scene/project opened and online). Green lines depict WebSockets connections (two-way communication necessary), blue are REST APIs, and for red, an implementer is free to choose appropriate technology.}
  \label{fig:block-diagram}
\end{figure}

The framework is divided into a set of services (backend) and a user interface (frontend). The main service of the system is \textbf{ARServer}, which acts as a central point for user interfaces and mediates communication with other services (see \cref{fig:block-diagram}).

So far, only one implementation of a user interface has been developed, a tablet-based app providing full functionality. However, the intention is to allow the involvement of several simpler, complementary interfaces providing only some aspect of functionality, e.g., RGB LED strip indicating system status or a hand tracking-based interface for controlling a robot. Therefore, the server must be able to deal with multiple connected interfaces, even in single-user scenarios. 

Interfaces are connected to ARServer through WebSockets, which allows bidirectional communication. The ARServer holds the system state, while interfaces can manipulate it using a set of RPCs and receive notifications on changes. It is assumed that each workplace runs a separate instance of ARServer (and all other related services), and therefore, the server maintains only one session for all users: if one user opens a project, the same project is shown to other connected users as well. To support efficient and safe collaboration between users, a locking mechanism prevents multiple users from manipulating the same element (e.g., controlling the robot).

The ARServer also serves as a proxy between Python code and the AR environment, which is code-agnostic. It analyzes the code of object types to extract available actions and their parameters and creates JSON metadata available to user interfaces. The code analysis takes advantage of PEP 484 type hints\footnote{\href{https://www.python.org/dev/peps/pep-0484/}{python.org/dev/peps/pep-0484/}} to extract, e.g., parameter types and matching nodes of Abstract Syntax Tree (AST)\footnote{\href{https://docs.python.org/3.8/library/ast.html}{docs.python.org/3.8/library/ast.html}} to, e.g., inspect value ranges, that are defined using assertions or check if a method (feature) is implemented.

The scene or project opened within the server could be either offline or online. In the online state, instances for all objects are created, meaning that, e.g., a connection to a robot is made. In an online state, robots could be manipulated, and any action instance added to a project may be executed, simplifying the programming and debugging process. However, it is also possible to work offline, where just functionalities such as controlling a robot are not available. Moreover, in the offline mode, the robot and other relevant machines do not have to be connected; therefore, the operator may prepare the base program in advance without needing the actual robot.

\textbf{Project Service} provides persistent storage for workplace-relevant data: scenes, projects, object types, models, etc.

\textbf{Scene Service}, used, e.g., in cases where underlying implementation is based on ROS, is responsible for managing collision objects.

The \textbf{Build Service} creates for a given project a self-contained executable package. The logic could be defined within the AR environment or provided in a standalone file. When generating logic from its JSON representation, it is first assembled in the form of AST and then compiled into Python code. Moreover, a set of supplementary classes, e.g., simplifying work with action points, are generated.

\textbf{Execution Service} manages execution packages created by the Build Service. The most important functionality is running the package when the service streams events regarding the execution state (e.g., which action with what parameters is being executed) to ARServer. The execution can also be paused or resumed when needed.

\textbf{Calibration Service} provides a method to perform camera pose estimation based on ArUco marker detection \cite{romero2018speeded}. The service is configured with IDs, poses, and sizes of available markers. When estimation is requested, markers are detected in the provided image, respective camera poses are computed, and then all poses are averaged using a camera-marker distance and camera-marker orientation as weights to produce the final 6D pose. For averaging quaternions, a method from \cite{markley2007averaging} is used. This estimation can be either used by user interfaces where, e.g., an AR visualization needs to be globally anchored, or ARServer could use it to update the camera's pose in a scene. Another method of the service may be used to adjust the pose of the robot using an RGBD camera. The robot model in a configuration corresponding to the actual robot state is rendered from the point of view of a camera in a robot's current position within the scene, which therefore serves as an initial guess. The virtual camera is used to generate a point cloud, which is then registered using a robust ICP (TukeyLoss kernel) with the point cloud from the real RGBD camera observing the scene (1024 frames averaged), which has been filtered to contain only close surroundings  around the robot in its initial pose. If the precision of such calibration is not enough for the task, more precise methods must be used, and the pose of the robot or camera can be entered manually.


\label{subsec:program-representation}
\subsection{Program Representation}  


One of the framework's goals is to support collaboration between non-programmers who prefer creating programs visually and programmers who prefer to work with code. Because of this, there are two language representations. For visual programming, there is an intermediate program representation based on JSON format, which is language agnostic, easy to serialize, supports common programming techniques (cycles, conditions, variables), allows flexible parameter specification, and is easily manipulable from user interfaces. For execution, the intermediate format is translated into Python, which is currently the most popular scripting language\footnote{According to PYPL Index for August 2022.}. The same language is also used to implement object types, through which a new device can be integrated into the system. This also allows a use case when a non-expert user creates the program visually, and the result is adjusted by an expert programmer. The form of Python code was designed with the possibility of transferring the code back into the intermediate format. However, this was not implemented yet.

\begin{figure}
  \includegraphics[width=0.98\textwidth]{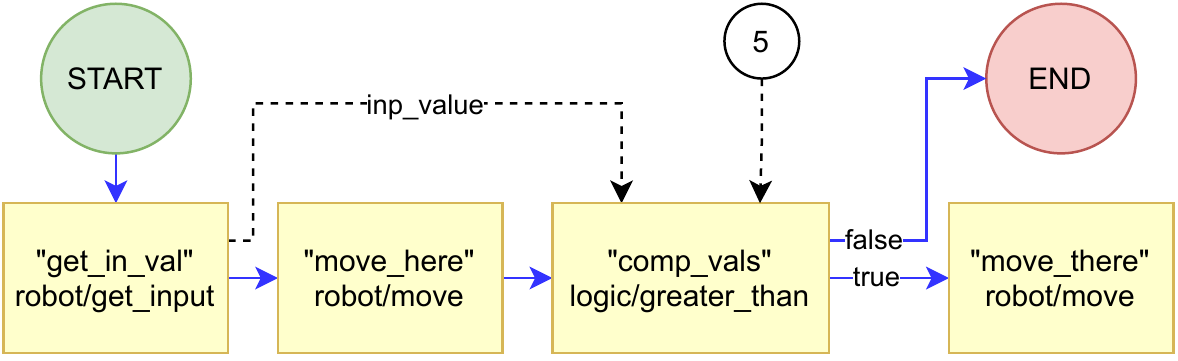}
  \centering
 \caption{The logical structure of an example program. Yellow boxes are actions (text in quotes is the user-entered name for the instance of action, below is an object and the corresponding method), blue lines denote logical connections (flow of the program), and black dashed lines denote data connections. The example shows how previous results can be used as parameters of a subsequent action and how logical flow can be branched based on a numerable value (boolean in this case).}
  \label{fig:program-example}
\end{figure}

The structure of the JSON format is as follows. Within a project, there might be $\interval{0}{n}$ action points, where each might contain $\interval{0}{n}$ actions. Each action is assigned a UID, unique human-readable name, type (scene object UID and corresponding underlying method/action), and $\interval{0}{n}$ parameters (corresponding to parameters of the method). Action parameters can be given as literal or referenced to either a project variable (constant shared by multiple actions) or a previous result (return value of precedent action). On the project level, there is an array of objects defining logical structure (visualized as blue lines, see \cref{fig:ui}), each containing UID of source and target action and optional condition. Actions together with those linkages then form a directed acyclic graph, where the loops are forbidden at an application level. Without a condition, two actions could be connected only with one logic linkage. Conditions are meant to achieve simple branching for numerable types such as boolean, enumerations, and integers. E.g., for branching according to a boolean value, two logical linkages are added, one for \textit{true} and the other for \textit{false} (see \cref{fig:program-example} and \cref{lst:program-example}). Any other type of condition has to be implemented in the form of action, for instance, \textit{greater\_than(float1, float2)} returning a boolean value. Also, loops are not part of the format definition and have to be implemented in the form of custom actions. This restriction keeps the intermediate format simple and, at the same time, allows integrators to provide a customized set of actions to their end-users.

\lstset{language=python,
        basicstyle=\ttfamily\scriptsize,
        breaklines=true
          }
\begin{lstlisting}[caption={An example of generated Python code. Parameter \texttt{an} denotes action name, which is human-readable counterpart to action UID.},captionpos=b, ,label={lst:program-example}]
inp_value = robot.get_input(an="get_in_val")
robot.move(an="move_here")
comp_res = logic.greater_than(inp_value, 5, an="comp")

if comp_res == True:
  robot.move(an="move_there")
\end{lstlisting}


\label{subsec:user-interaction}
\subsection{User Interaction}



A working prototype, based on the concept presented by Kapinus et al. \cite{kapinus2019spatially} was implemented and iteratively tested and improved in cooperation with the project partner. The application's design was modified to support the two-handed operation of the tablet and control of interface elements using the user's thumbs, to lower the fatigue of arms and hands.  

The primary concept of our tablet user interface deals with the fact that most robotic programs interact with a real environment in some manner. An operator, using our user interface, can annotate the environment in a simplified way and subsequently design programs' logic. Thanks to the utilization of AR, this can happen within the task space, and therefore mental demands are lowered \cite{huang2020contextual, gadre2019end}.

Our user interface offers several graphical elements to precisely annotate specific places in the environment (action points). These places may later be used as spatial anchors for elements representing specific actions. Visual elements representing actions (also known as \textit{pucks} in our GUI) are therefore located at the place where the action will be executed, which improves users' comprehension of spatial relations within the program.

\begin{figure}
  \includegraphics[width=0.78\textwidth]{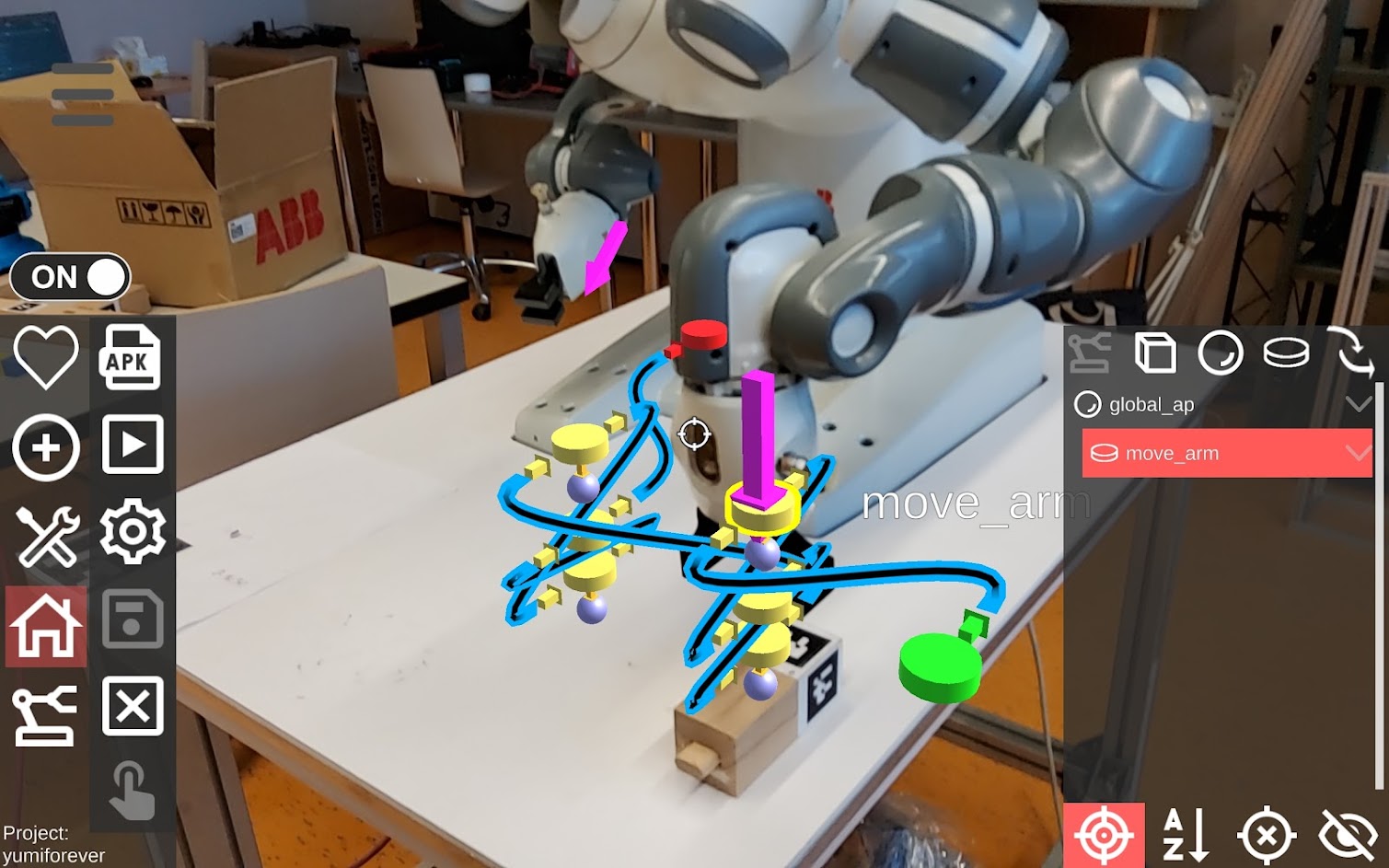}
  \centering
  \caption{The application screen with a tool menu (left), selector menu (right), non-interactive 3D scene (center), with crosshair, action points (violet), actions (yellow, green for program loop start, red for its end), logical connections (black/blue) and robot end effectors (magenta). The visualized program realizes a simple pick-and-place task consisting of low-level actions.}
  \label{fig:ui}
\end{figure}

The whole interface consists of three main parts: \textit{crosshair} in the middle of the screen, \textit{main menu} on the left side of the screen, and \textit{tool context menu} on the right side of the screen. The \textit{crosshair} is used to select virtual objects by the physical pointing of the tablet. \textit{The tool selection menu} shows actions available for a currently selected object (e.g. duplicate object, transform object, etc.) and the \textit{tool context menu} serves as a sub-menu for currently performed operations (e.g. move / rotate tools when \textit{action object} is being manipulated). See \cref{fig:ui}.

\label{sec:evaluation}
\section{Evaluation}

During development, the system and its user interface were evaluated multiple times using different methods and continuously refined to provide a plausible user experience and fulfill use case needs. Some of the presented evaluations focused on certain aspects of the problem were already published, and here, we only briefly report their results to provide a complete overview summarizing the framework's evolution.

\subsection{Interface Concept}

A high-fidelity initial prototype of a user interface was implemented and evaluated in a qualitative experiment ($n=7$) by Kapinus et al. \cite{kapinus2020end}, which focused on usability, mental workload, and user experience. Participants were asked to create a simple program using several highly abstract actions. The task consisted of 21 steps that had to be done.


Regarding subjective metrics, the user interface was ranked A using SUS~\cite{brooke1996sus} (90-95th percentile), and it was overall rated Excellent in all UEQ~\cite{schrep2015ueq} categories, i.e., Attractiveness (mean score 1.93, SD=0.58), Pragmatic attributes (mean score 2.26, SD=0.28), and Hedonic attributes (mean score 1.86, SD=0.72). Overall, NASA-TLX~\cite{hart1988nasatlx} score was 27.38 (SD=9.41), making it lower than 80~\% of studies analyzed in \cite{grier2016}.

From the objective point of view, the main finding was that all participants could successfully complete the task in a reasonable time (average completion time was 527~s, SD=130~s). No fundamental problem was discovered, although some minor issues were observed or reported by participants.

Most participants complained about the design of \textit{pucks} (visual elements representing actions), mainly about their size, appearance, and placement strategy. Based on that, the design was altered, and because they are now placed above action points, the position of the puck could now be changed by the user if the default position is not suitable. 

The other important issue was related to the selection of virtual objects. In the prototype, the object was selected by touching it on the screen, causing the user to hold the device with only one hand, so they could use the other hand to touch the object. This caused hand and arm fatigue. Moreover, the selection was sometimes difficult, especially in a cluttered environment. To cope with it, the design of the interface was changed to allow controlling all elements by a user's thumbs when holding the device with both hands in landscape mode, which also demanded the utilization of some indirect selection method.

Only two participants found out that they could benefit from the active movement of the mobile device inside the scene. One participant explicitly stated that they wanted to stay in one spot and zoom inside the scene. To solve this issue, a non-immersive VR mode was introduced, which allows users to change the position of a virtual camera freely and, therefore, to see and interact with the workplace from any angle and distance.

\subsection{Virtual Object Selection}


As the initial experiment with the prototype of the interface has shown, the direct selection of virtual objects on the screen was problematic in some cases (high density of objects, occlusions, similar objects), and the necessity to hold the device with one hand and touch the screen with the other lead to ergonomic issues. Therefore, the design of the application was changed in a way that all interactable elements are within reach of the user's thumbs, and we implemented and compared two indirect selection methods, which were described and preliminarily evaluated in \cite{kapinus2022improved}. Both of them work with spatially clustered objects. One method is based on a spatially anchored hierarchy menu, and the other utilizes a crosshair and a side menu that shows candidate objects according to a custom-developed metric. Compared with the direct touch method, the results indicate that indirect methods might lead to better precision and improved confidence in selecting the intended object, however, at the expense of worse task performance. The methods will be further developed and evaluated. However, based on the experiment's results, we chose the method with a side menu (the \textit{selector menu}, see \cref{fig:ui}) as it seems to provide better performance (time to select an object), lower task load (TLX) and similar success rate as the hierarchical menu.

\subsection{Non-immersive VR mode}

When working with AR, there is often a need to move closer to distinguish or inspect virtual objects or, on the contrary, to move further to see the whole scene, which is amplified by the limited field of view of handheld devices and HMDs. Regarding handheld devices, it is also often necessary to see the scene from different angles, to correctly judge the placement of the objects, which is caused by a lack of depth perception due to monoscopic display. Moreover, within industrial environments, there is typically limited floor space, safety curtains, fences around robot cells, etc. These constraints might make viewing the workplace from certain poses physically challenging or even impossible. Therefore we proposed and evaluated an approach allowing temporal switching from AR to a non-immersive VR \cite{bambusek2022handheld}. In VR, the application shows a 3D model of the workplace, and the viewpoint is controlled by device motion in conjunction with on-screen joysticks, with non-linear sensitivity. The conducted experiment ($n=20$), based on the object alignment task, revealed that self-reported physical demands are significantly lower when users are allowed to switch between AR and VR arbitrarily. The usefulness of the VR mode was rated as high, and during the task, users spent 70~\% of the time within it. Observations of users' behavior have revealed that the VR mode was often used to get an overview of the workspace, to find an occluded object, or to avoid an uncomfortable position.

\subsection{Multi-user Collaboration}

To evaluate the collaborative aspects of the framework, a small-scale user experiment ($n=3$) was conducted in a lab-like environment. The experiment was focused on functionality; however, it also served as the very first usability evaluation. The workplace consisted of two robots (Dobot M1 and Magician), a conveyor belt, and several collision objects.

The task was to set up the workplace collaboratively and create a simple program for moving cubes from one robot to the other and back using the conveyor belt. In the setup phase, each participant added one action object (a robot or the belt) to the scene and positioned it. Subsequently, a project was created, and each participant created several action points (using hand-teaching and visual positioning tools) and related actions. Finally, the logical connections defining the program flow were added, and the project was executed. The participants worked on separate sub-tasks most of the time but shared the same workspace. Moreover, they had to collaborate to connect all sub-tasks into a working program successfully.

From the technical perspective, the user experience during the collaboration was smooth, and user interfaces were kept synchronized properly. Regarding usability, although users communicated verbally during the experiment, they also appreciated visual indication of which object is being used (locked) by someone else. Based on the course of the experiment, collaborative programming seems to be a viable approach, and we will investigate it more deeply in the forthcoming research. In this case, the task was done by three users in approximately 15 minutes. It would be interesting to determine the relation between task time and the number of collaborating users on a significantly more complex task.
\begin{figure*}
\centering
\subfloat[During hand teaching, the robot is locked by the user and therefore unavailable for others, which is indicated within the 3D scene.]{\includegraphics[height=0.335\textwidth]{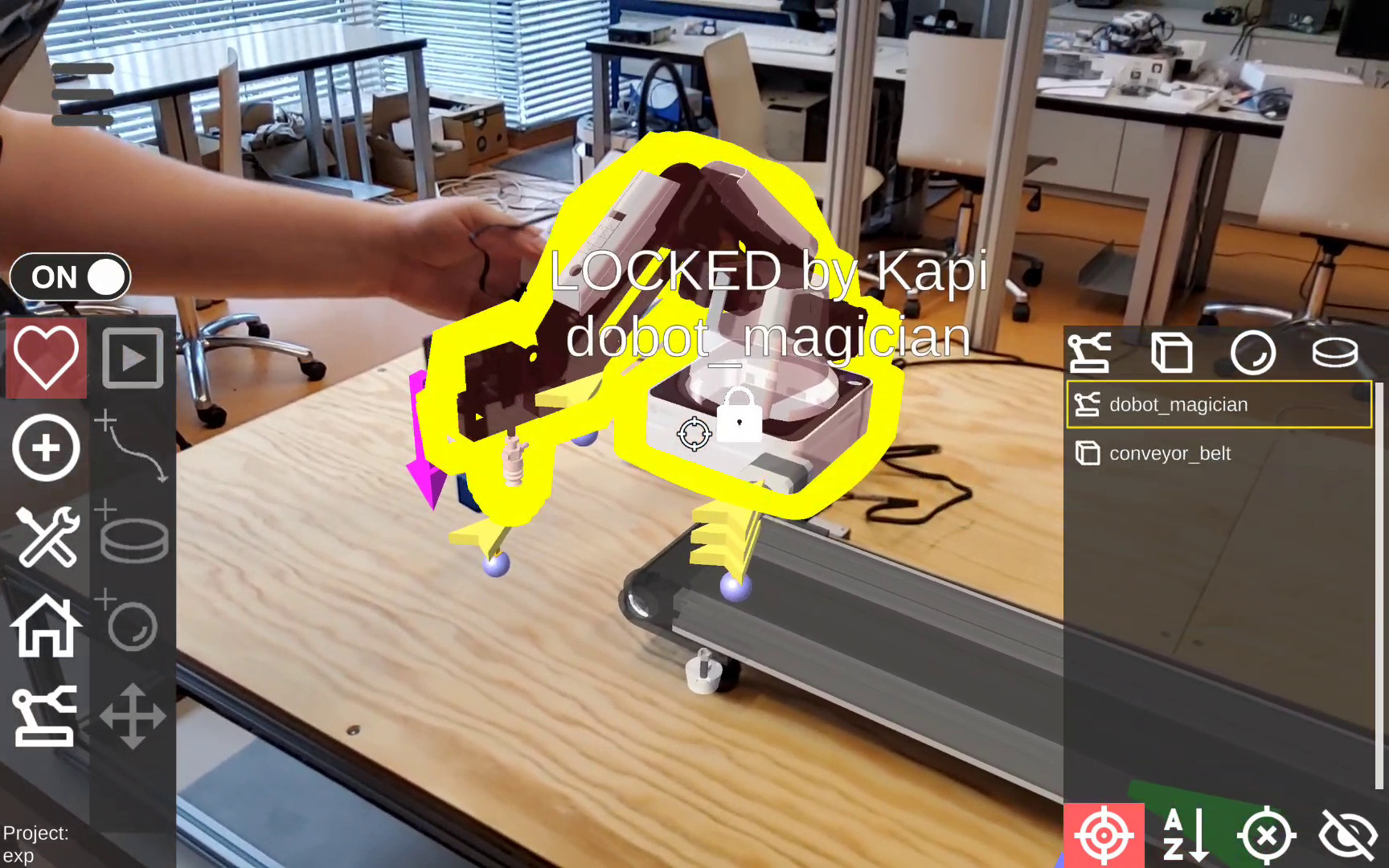}}
\hfill
\subfloat[Multiple users collaborating on the task of moving boxes between robots using a conveyor belt.]{\includegraphics[height=0.335\textwidth]{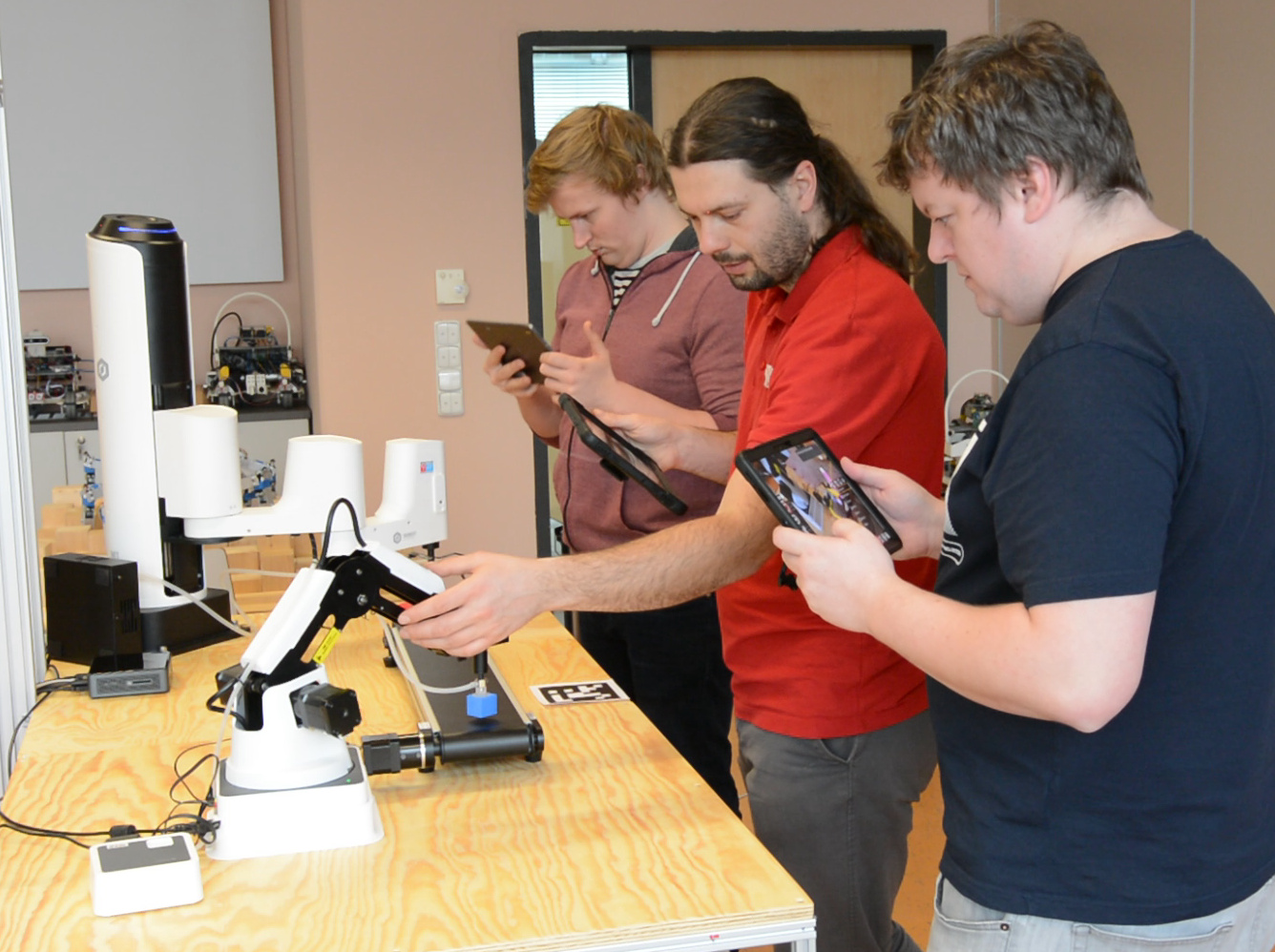}}

\caption{Technically oriented evaluation of the collaborative-related functionality.}
\label{fig:collab}

\end{figure*}


\subsection{Iterative Testing and Refinement}

The whole system was created in close cooperation with our industrial partner, who has over 15 years of experience in automation. During the development, the design of the user interface and key parts of the system were discussed in detail with the system integrator and potential end-users. 

Besides individual testing, there were several integration meetings where the whole task programming process was tested with the real production-like robotic cell to maximize simplicity and comprehensibility for the end-users. This helped us better understand real-world scenario difficulties, which are not obvious during in-lab testing, e.g., object selection in heavily cluttered environments (e.g., caged robotic cell), dealing with safety precautions, etc. 

As a result, several improvements were added to the user interface, for example, a better ray-casting strategy for virtual object selection, enabling a user to disable several objects (e.g., virtual safety walls), which have to be part of the scene, but once they are created, are barely used anymore.

Additionally, an unstructured interview with the system integrator representative was organized. They were asked to state their opinion on the framework being developed from the commercial perspective. The following key advantages were claimed: 


\begin{itemize}
    \item Clear visualization of position and distribution of individual actions in space.
    \item Ease and speed of programming for smaller-scale applications.
    \item Possibility of visual composition of the scene -- collision objects, positional relationships of individual elements.
    \item Simple and useful controls for the robot -- stepping, end effector alignment, integration of hand teaching.
\end{itemize}


And the following limitations were pointed out:

\begin{itemize}
    \item Visual clutter for large-scale applications -- too much graphical information.
    \item Ergonomically demanding method requiring the creation of the entire application in a standing position while holding a tablet.
\end{itemize}

Regarding large-scale applications, it was suggested to define categories of actions and to distinguish them, e.g., by color-coding or by icons. Moreover, the visual clutter may be reduced by different techniques, such as implementing, e.g., level of detail \cite{Tatzgern2016}, or by implementing more complex actions on a higher level of abstraction, which will reduce the number of individual actions needed to realize a given task. The ergonomy of use may be improved by the proper holding of the device (needs to be covered during training) \cite{Veas2008} and also by the already described VR mode, which allows users to temporarily switch from AR to VR to reach physically unreachable poses, zoom in, or to work while sitting.

\subsection{Expert Review}


When the system, interface design, and features were mostly stable, an expert review was conducted to eliminate the most significant user experience problems and validate the overall concept of the system at the \textit{End User} level (category). Three reviews were obtained.

The first reviewer (\(R_A\)) is an experienced software tester. The review was performed at the testing site of the project partner with an Aubo i5 and one custom-built two-axis robot. The second reviewer (\(R_B\)) is a 3D data visualization and processing specialist with experience in human-robot interaction. The third one (\(R_C\)) is an expert in cyber-physical systems, computer graphics, user interface design, and evaluation.

Reviews by \(R_B\) and \(R_C\) were performed at the university robotics laboratory with an ABB YuMi robot. Reviewers \(R_A\) and \(R_B\) used the same version of the interface, while \(R_C\) used a slightly updated one that was available at the moment. Each session lasted approximately one and a half hours. The Samsung Galaxy Tab S6 with a protective cover was used. A reviewer was given a technical document describing the solution in advance and then briefly introduced to the usage of the interface. Then, they went through the core functionality while commenting on their findings. The comments were recorded and, after the experiment, processed into a review protocol. The reviewer was then asked to verify the protocol, briefly comment on each issue, and assign severity on a scale of $\interval{1}{5}$.

\begin{figure}
  \includegraphics[width=0.88\textwidth]{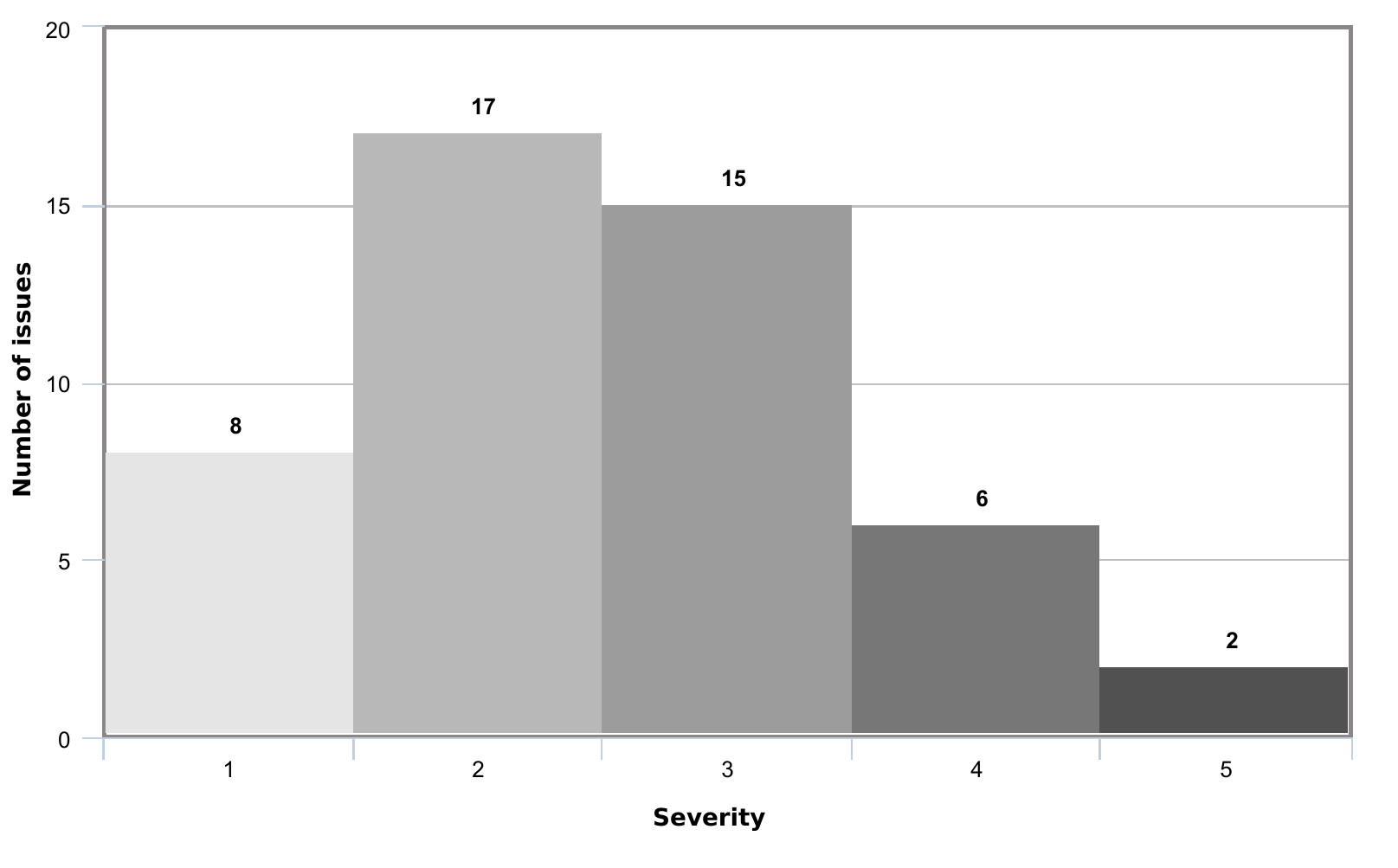}
  \centering
  \caption{Histogram of all reported issues clustered by their severity.}
  \label{fig:histogram}
\end{figure}

39 unique issues (48 in total) of different severity (see \cref{fig:histogram}) were identified by all reviewers. Most of them were ranked with low severity, dealing mostly with some minor user interface usability issues, such as difficult number input (\(R_A\)) caused by the default Android keyboard, the unclear icon for a favorite group of actions (\(R_B\)), or issue with main menu actions grouping, forcing the reviewer to navigate through the menu to locate required action (\(R_C\)). The most severe issues are shown in \cref{tab:most-severe-issues}, with the reviewer's self-reported severity and a brief suggestion.

The collected feedback was categorized into the following groups:
\begin{itemize}
    \item \textbf{Control} (12 unique issues, 14 total) -- issues related to user interface control.
    
    \item \textbf{HUD design} (13 unique issues, 17 in total) -- problems with the user interface itself -- icons, menu design, etc.
   
    \item \textbf{System Status} (3 unique issues, 4 in total) -- related to notifications and system state visualization. 
    
    \item \textbf{Visualization} (11 unique issues, 13 in total) -- visualization of 3D scene content (Action Points, Actions, etc.).
    
\end{itemize}

\renewcommand\tabularxcolumn[1]{m{#1}}

\begin{table*}[htbp]
\centering
\caption{The most severe issues (rated 4 or 5) and suggestions on how to mitigate them as reported by the reviewers.}
\begin{tabularx}{\textwidth}{lXcXc}
\toprule
\multicolumn{1}{c}{} & \multicolumn{1}{c}{\textbf{Issue}} & \multicolumn{1}{c}{\textbf{Severity}} & \multicolumn{1}{c}{\textbf{Suggestion}} & \multicolumn{1}{c}{\textbf{Category}}\\ \cmidrule(lr){2-5}
\multirow{7}{*}{\(R_A\)}       & Too many buttons and time-consuming determination of their meaning.  & 5 & Implement a guide helping new users to get comfortable with application usage. & Documentation \\ \cmidrule(lr){2-5}
                               & System status not visible in some cases (e.g., in case of long-lasting processes as calibration). & 4 & Add persistent notifications. & System Status \\ \cmidrule(lr){2-5}
                               & Inability to aim objects through another object. & 4 & Add the possibility to disable an obstructing object temporarily. & Control \\ \cmidrule(lr){2-5}
                               & Physical demands when working longer than 30 minutes. & 4 & Encourage both hand-holding, suggest rest period. & Control \\ \cmidrule(lr){1-5}
\multirow{2}{*}{\(R_B\)}       
                               & A skill and a lot of physical movement are needed to judge the position of virtual objects. & 5 & Add shadows to the virtual objects to improve depth perception. & Visualisation \\ \cmidrule(lr){2-5}
                               & Complicated flow to add an action. & 4 & Allow to add an action without adding an action point first. & Control \\ \cmidrule(lr){1-5}
\multirow{2}{*}{\(R_C\)}       & Reachability of \textit{action points} by selected robot is not visualised. & 4 & Add some indication of reachability.            &  Visualisation    \\ \cmidrule(lr){2-5} 
                               & Difficult orientation in more complicated programs & 4 & Use different connections for different logic flow phases.          & Visualisation       \\ \bottomrule
\end{tabularx}
\label{tab:most-severe-issues}
\end{table*}


The \(R_A\) pointed out that there are too many icon buttons, and their purpose is not always very clear at first sight. The only way to understand individual icons is to hold a finger over them until a tooltip is shown. This issue will be solved by providing proper application documentation and onboarding mode, which will guide the novice user through the application. 


The reviewer also pointed out that it was quite physically demanding to hold the tablet for a longer period of time, and it was necessary to take rest periods regularly. It could be partially caused by improper holding of the device, where the \(R_A\) held it by one hand on the left side of the screen for some time at the beginning of the review, which caused fatigue to the arm. This issue could be addressed in the onboarding mode of the application, where proper holding should be demonstrated and rest sessions suggested.


The testing site of the project partner contained virtual walls around the robotic cell assembled from dummy action objects. This complicated object selection because these safety walls were the first objects to be hit by a ray from crosshair, making virtual objects inside the cell virtually impossible to select by aiming. This issue was already solved by enabling users to put selected virtual objects on the blocklist, thus excluding them from the selection process.

On the other hand, the \(R_A\) liked the visualization of the logical flow of the program, which helped them both understand the meaning of the edited program and change the program's behavior. Moreover, the reviewer appreciated the simplicity of robot stepping available directly from the programming application without using the dedicated teach pendant. The reviewer explicitly mentioned that the ability to align the robot's end-effector with the underlying table was crucial for the fast navigation of the robot.

The main issue for the second reviewer, the \(R_B\), was related to depth sensing. The reviewer had problems understanding where the manipulated object (e.g., action point) is in the real world, and they had to walk around heavily to see its position from multiple angles. They pointed out that it was probably caused by the lack of generated shadows, which usually helps people to sense depth. As a (partial) solution, we have enabled shadows and light estimation in our application. To support the user's knowledge of object position, a kind of projection in the plane of the table, with information on the height above the table (or nearest object below), could be incorporated.

The reviewer also stated that the flow of action definition is unnecessarily complex and difficult, meaning that the user has to insert an action point and then assign an action to it. They suggested that simplification of this process would be a significant improvement. Among others, this particular issue is currently being more deeply investigated in a separate study, which will be published later.  


Last but not least reviewer, the \(R_C\), stated that it makes sense for them to define actions inside the real environment. On the other hand, they were worried that it would be difficult to understand a more complex program where many actions together with conditional execution would be incorporated. The reviewer suggested that selected logical flow parts should be differentiated by color or shape. 

Another important issue for the \(R_C\) was that the robot's reachability of action points and executability of actions was not visualized in any way. For actions, a mark could be easily added to indicate whether or not the action could be executed. Regarding the reachability of action points, the possibility of having more than one robot in the scene must be considered. Moreover, any action point could potentially have several orientations, each needing to be evaluated separately. 

In summary, the reviewers approved the overall concept of the framework as suitable for end-users. As they tested the framework through the user interface, naturally, most findings were related to its usability, where several relevant suggestions were collected.





\section{Lessons Learnt}


The problem of designing a framework that should act as a central point of integration for arbitrary devices and its user interface should provide great usability for end-users while serving a high amount of dynamically loaded functionality is complex, and there certainly exist different ways to approach it. From our experience, it is crucial to have a realistic use case and then define what type of users will interact with the framework and its user interface in what ways. We see the iterative design process for both the backend architecture, API, and user interface design, as a key factor determining success. We would also suggest making smaller and specifically focused experiments as we did, rather than one all-embracing, because it allows faster iteration and provides better interpretable results. 


The iterative design process led us to a state where the user interface is minimalistic and optimized for best performance when used regularly, for several hours a day. The drawback is that some of its aspects as the indirect selection of objects are slightly unintuitive for novel users. Moreover, there is specific terminology, etc. Therefore, a need for some form of initial training is inevitable. The training, when executed efficiently (video, in-application guide), is, in our opinion, worth the improvement of performance during regular operation.




One of the main problems was overcoming the imprecise registration of AR visualization to the real environment. We chose to rely on work with relative coordinates, where the precise coordinate is obtained through a robot (usually the most precise device at the workplace), and then further points are added relative to it. Points are manipulated by visual tools providing virtually unlimited precision. In practice, it turned out that it is highly useful to let users navigate the robot to any point, then step the robot and update the point. 

From the practical point of view, it proved useful to organize the source codes of the backend (17 Python packages at the moment) in a monorepo, utilizing the Pants build system\footnote{\href{https://www.pantsbuild.org/}{https://www.pantsbuild.org/}}. Especially in the early phases of development, monorepo makes major refactoring easier. Moreover, it significantly reduces necessary effort related to maintenance as there is only one configuration of continuous integration, etc. Although using tests may sound obvious, it is often skipped for academic software. The adoption of continuous integration and implementation of different levels of tests (unit, integration) definitely took some resources, but finally, it saved us a lot of valuable time and made us confident when performing changes to the code base.

\label{sec:conclusions}
\section{Conclusions}

The paper presented the ARCOR2 framework, which allows end-users to program not only robots but whole complex workplaces or production lines consisting of multiple robots and other arbitrary machines. The framework was designed, developed, and iteratively tested in close cooperation with the industrial partner, who provided the realistic use case, testing site, and valuable feedback. One of the main advantages of the solution is that support for any device or service could be added as a plugin. The visual programming in the AR interface allows specifying not only program steps and their parameters but also logical flow, including conditions. Kinesthetic teaching is utilized to obtain precise positions; however, its usage is minimized to limit users' fatigue only to get reference points, and then other necessary points are manipulated by graphical tools relative to the robot-originating points. This way, we also cope with the inaccuracy of AR registration to the real world. The framework also supports multiple users working simultaneously, which can be useful, for instance, for large-scale workplaces or during training.

Many rounds of internal testing were performed, focused both on the user interface as well as on the API of the framework, making sure that it is usable from both end-user and expert-user points of view. The role of expert users is mainly to develop the integration of devices and services and prepare necessary functionality on a task-appropriate level of abstraction. The end-users role is to program the task, but as the visual representation is compiled into a source code, expert users can get easily involved even in this stage when needed. Despite internal testing, the initial concept of the user interface was evaluated in a user study. After that, the interface design was changed, e.g., to allow controlling all UI elements by users' thumbs and thus reduce fatigue from holding a tablet in one hand. After the functionality was stabilized, an expert review was carried out, and the results are presented in this paper. Its main outcomes are validation of the overall concept as well as several identified points for further development, such as the necessity to implement an onboarding mode to support novice users, to improve depth perception by providing additional cues, and to simplify the flow for adding actions.

In summary, the framework, when adopted by a system integrator and its customer, allows efficient collaboration between professional (robot) programmers and end-users, who are domain experts. By allowing end-users to set up a workplace and create or adapt the program, the high flexibility needed for SMEs is achieved. Moreover, as SMEs can perform program modifications/adjustments on their own, expenses are reduced.


The gained experience also helped us to answer our long-term research questions at least partially. Within the context of visual programming, it turned out that situated interaction could be realized using AR and handheld devices when supported by specifically designed interfaces (Q1-2). The solution we chose to overcome imprecise registration of AR visualization was to utilize the robot's precision and then move virtual objects relatively to precisely obtained positions. Then, visualization might be, for instance, slightly shifted or not fully stable, but the spatial relations between objects are kept, and users can manipulate objects with arbitrary precision (Q3). However, it has to be clearly communicated to users that such imperfection might occur and that it has no impact on the robot's precision. From our experience, a low level of abstraction makes visual representation incomprehensible, and therefore basic program steps must be merged into more complex, parameterizable skill-like ones (Q4).

\subsection{Future Work}

There is ongoing work on comparing ARCOR2 with an existing, commercially available solution for the simplified programming of robots. The paper under preparation will also contain an in-detail description of the user interface, which was here described only briefly. At the same time, we are planning an out-of-the-lab long-term study that will start once the framework is deployed at a PCB testing facility.

In the current work, the expert review was performed on the \textit{End User} level. However, the system will be used by \textit{Expert} level users as well. In this regard, we will formally evaluate their user experience with related development, integration of new devices, and debugging errors.

Moreover, collaborative programming by multiple end-users deserves to be investigated more deeply, including evaluating the effectiveness of such an approach for larger workplaces. There is also a plan to implement decompilation of Python code into visual representation allowing bidirectional synchronization between the AR environment and source code (for this to be possible, a source code would have to follow established conventions), which would, in turn, allow evaluation of collaboration between experts and end-users.

We also plan to port and adapt the user interface to other platforms, e.g., HoloLens 2, and to add support for advanced programming constructs like functions or visually defined object actions allowing code reuse. It will also be highly interesting to investigate how the framework could be applied to other use cases than PCB testing, which predominantly consists of picking and placing parts.

Finally, during the development of the framework, many general questions related to handheld AR arose. For instance, what are optimal cues to compensate for the limited depth perception, or how to motivate users (who tend to be rather still during AR usage) to take advantage of free movement? These questions represent motivation for further long-term research.

\vspace{7pt}

\noindent\small{\textbf{Code availability}} The source codes are available for download through GitHub.

\vspace{7pt}

\noindent\small{\textbf{Data availability}} The datasets generated during the current study are not publicly available due to the participants' privacy but are available from the corresponding author on reasonable request.

\vspace{7pt}

\noindent\small{\textbf{Funding}} This work has been performed in the scope of the 5G-ERA European
Research Project and has been supported by the Commission of the European Communities
(Grant Agreement No. 101016681).

\section*{Declaration}

\noindent\small{\textbf{Author Contributions}} All authors contributed to the study conception and design. Prototypes and experiments were prepared and performed by Michal Kapinus, Zdeněk Materna, and Daniel Bambušek. The first draft of the manuscript was written by Zdeněk Materna and all authors commented on previous versions of the manuscript. All authors read and approved the final manuscript.

\vspace{7pt}

\noindent\small{\textbf{Consent of participate}} Available, signed by each participant.

\vspace{7pt}

\noindent\small{\textbf{Consent to publish}} The authors affirm that human research participants provided informed consent for publication of the images in Figures 6a and 6b.

\vspace{7pt}

\noindent\small{\textbf{Conflict of interest}} There are no conflicts of interest/competing interests.

\vspace{7pt}

\noindent\small{\textbf{Ethical statement}} The manuscript complies with the Ethical Rules
applicable for this journal as stated in the Instructions for Authors of
the Journal of Intelligent \& Robotics Systems.


\bibliography{literature}


\begin{thebibliography}{31}
\ifx \bisbn   \undefined \def \bisbn  #1{ISBN #1}\fi
\ifx \binits  \undefined \def \binits#1{#1}\fi
\ifx \bauthor  \undefined \def \bauthor#1{#1}\fi
\ifx \batitle  \undefined \def \batitle#1{#1}\fi
\ifx \bjtitle  \undefined \def \bjtitle#1{#1}\fi
\ifx \bvolume  \undefined \def \bvolume#1{\textbf{#1}}\fi
\ifx \byear  \undefined \def \byear#1{#1}\fi
\ifx \bissue  \undefined \def \bissue#1{#1}\fi
\ifx \bfpage  \undefined \def \bfpage#1{#1}\fi
\ifx \blpage  \undefined \def \blpage #1{#1}\fi
\ifx \burl  \undefined \def \burl#1{\textsf{#1}}\fi
\ifx \doiurl  \undefined \def \doiurl#1{\url{https://doi.org/#1}}\fi
\ifx \betal  \undefined \def \betal{\textit{et al.}}\fi
\ifx \binstitute  \undefined \def \binstitute#1{#1}\fi
\ifx \binstitutionaled  \undefined \def \binstitutionaled#1{#1}\fi
\ifx \bctitle  \undefined \def \bctitle#1{#1}\fi
\ifx \beditor  \undefined \def \beditor#1{#1}\fi
\ifx \bpublisher  \undefined \def \bpublisher#1{#1}\fi
\ifx \bbtitle  \undefined \def \bbtitle#1{#1}\fi
\ifx \bedition  \undefined \def \bedition#1{#1}\fi
\ifx \bseriesno  \undefined \def \bseriesno#1{#1}\fi
\ifx \blocation  \undefined \def \blocation#1{#1}\fi
\ifx \bsertitle  \undefined \def \bsertitle#1{#1}\fi
\ifx \bsnm \undefined \def \bsnm#1{#1}\fi
\ifx \bsuffix \undefined \def \bsuffix#1{#1}\fi
\ifx \bparticle \undefined \def \bparticle#1{#1}\fi
\ifx \barticle \undefined \def \barticle#1{#1}\fi
\bibcommenthead
\ifx \bconfdate \undefined \def \bconfdate #1{#1}\fi
\ifx \botherref \undefined \def \botherref #1{#1}\fi
\ifx \url \undefined \def \url#1{\textsf{#1}}\fi
\ifx \bchapter \undefined \def \bchapter#1{#1}\fi
\ifx \bbook \undefined \def \bbook#1{#1}\fi
\ifx \bcomment \undefined \def \bcomment#1{#1}\fi
\ifx \oauthor \undefined \def \oauthor#1{#1}\fi
\ifx \citeauthoryear \undefined \def \citeauthoryear#1{#1}\fi
\ifx \endbibitem  \undefined \def \endbibitem {}\fi
\ifx \bconflocation  \undefined \def \bconflocation#1{#1}\fi
\ifx \arxivurl  \undefined \def \arxivurl#1{\textsf{#1}}\fi
\csname PreBibitemsHook\endcsname

\bibitem{ionescu2019participatory}
\begin{barticle}
\bauthor{\bsnm{Ionescu}, \binits{T.B.}},
\bauthor{\bsnm{Schlund}, \binits{S.}}:
\batitle{A participatory programming model for democratizing cobot technology
  in public and industrial fablabs}.
\bjtitle{Procedia CIRP}
\bvolume{81},
\bfpage{93}--\blpage{98}
(\byear{2019})
\end{barticle}
\endbibitem

\bibitem{schmidbauer2020teaching}
\begin{barticle}
\bauthor{\bsnm{Schmidbauer}, \binits{C.}},
\bauthor{\bsnm{Komenda}, \binits{T.}},
\bauthor{\bsnm{Schlund}, \binits{S.}}:
\batitle{Teaching cobots in learning factories--user and usability-driven
  implications}.
\bjtitle{Procedia manufacturing}
\bvolume{45},
\bfpage{398}--\blpage{404}
(\byear{2020})
\end{barticle}
\endbibitem

\bibitem{ajaykumar2020user}
\begin{bchapter}
\bauthor{\bsnm{Ajaykumar}, \binits{G.}},
\bauthor{\bsnm{Huang}, \binits{C.-M.}}:
\bctitle{User needs and design opportunities in end-user robot programming}.
In: \bbtitle{Companion of the 2020 ACM/IEEE International Conference on
  Human-Robot Interaction},
pp. \bfpage{93}--\blpage{95}
(\byear{2020})
\end{bchapter}
\endbibitem

\bibitem{weiss2016first}
\begin{barticle}
\bauthor{\bsnm{Weiss}, \binits{A.}},
\bauthor{\bsnm{Huber}, \binits{A.}},
\bauthor{\bsnm{Minichberger}, \binits{J.}},
\bauthor{\bsnm{Ikeda}, \binits{M.}}:
\batitle{First application of robot teaching in an existing industry 4.0
  environment: Does it really work?}
\bjtitle{Societies}
\bvolume{6}(\bissue{3}),
\bfpage{20}
(\byear{2016})
\end{barticle}
\endbibitem

\bibitem{huang2020contextual}
\begin{bchapter}
\bauthor{\bsnm{Huang}, \binits{C.-M.}}:
\bctitle{Contextual programming of collaborative robots}.
In: \bbtitle{International Conference on Human-Computer Interaction},
pp. \bfpage{321}--\blpage{338}
(\byear{2020}).
\bcomment{Springer}
\end{bchapter}
\endbibitem

\bibitem{paxton2017costar}
\begin{bchapter}
\bauthor{\bsnm{Paxton}, \binits{C.}},
\bauthor{\bsnm{Hundt}, \binits{A.}},
\bauthor{\bsnm{Jonathan}, \binits{F.}},
\bauthor{\bsnm{Guerin}, \binits{K.}},
\bauthor{\bsnm{Hager}, \binits{G.D.}}:
\bctitle{Costar: Instructing collaborative robots with behavior trees and
  vision}.
In: \bbtitle{2017 IEEE International Conference on Robotics and Automation
  (ICRA)},
pp. \bfpage{564}--\blpage{571}
(\byear{2017}).
\bcomment{IEEE}
\end{bchapter}
\endbibitem

\bibitem{stenmark2017simplified}
\begin{bchapter}
\bauthor{\bsnm{Stenmark}, \binits{M.}},
\bauthor{\bsnm{Haage}, \binits{M.}},
\bauthor{\bsnm{Topp}, \binits{E.A.}}:
\bctitle{Simplified programming of re-usable skills on a safe industrial robot:
  Prototype and evaluation}.
In: \bbtitle{Proceedings of the 2017 ACM/IEEE International Conference on
  Human-Robot Interaction},
pp. \bfpage{463}--\blpage{472}
(\byear{2017})
\end{bchapter}
\endbibitem

\bibitem{gao2019pati}
\begin{bchapter}
\bauthor{\bsnm{Gao}, \binits{Y.}},
\bauthor{\bsnm{Huang}, \binits{C.-M.}}:
\bctitle{Pati: a projection-based augmented table-top interface for robot
  programming}.
In: \bbtitle{Proceedings of the 24th International Conference on Intelligent
  User Interfaces},
pp. \bfpage{345}--\blpage{355}
(\byear{2019})
\end{bchapter}
\endbibitem

\bibitem{mayr2021considerations}
\begin{bchapter}
\bauthor{\bsnm{Mayr-Dorn}, \binits{C.}},
\bauthor{\bsnm{Winterer}, \binits{M.}},
\bauthor{\bsnm{Salomon}, \binits{C.}},
\bauthor{\bsnm{Hohensinger}, \binits{D.}},
\bauthor{\bsnm{Ramler}, \binits{R.}}:
\bctitle{Considerations for using block-based languages for industrial robot
  programming-a case study}.
In: \bbtitle{2021 IEEE/ACM 3rd International Workshop on Robotics Software
  Engineering (RoSE)},
pp. \bfpage{5}--\blpage{12}
(\byear{2021}).
\bcomment{IEEE}
\end{bchapter}
\endbibitem

\bibitem{huang2017code3}
\begin{bchapter}
\bauthor{\bsnm{Huang}, \binits{J.}},
\bauthor{\bsnm{Cakmak}, \binits{M.}}:
\bctitle{Code3: A system for end-to-end programming of mobile manipulator
  robots for novices and experts}.
In: \bbtitle{2017 12th ACM/IEEE International Conference on Human-Robot
  Interaction (HRI},
pp. \bfpage{453}--\blpage{462}
(\byear{2017}).
\bcomment{IEEE}
\end{bchapter}
\endbibitem

\bibitem{yigitbas2021simplifying}
\begin{botherref}
\oauthor{\bsnm{Yigitbas}, \binits{E.}},
\oauthor{\bsnm{Jovanovikj}, \binits{I.}},
\oauthor{\bsnm{Engels}, \binits{G.}}:
Simplifying robot programming using augmented reality and end-user development.
arXiv preprint arXiv:2106.07944
(2021)
\end{botherref}
\endbibitem

\bibitem{thoo2021online}
\begin{botherref}
\oauthor{\bsnm{Thoo}, \binits{Y.J.}},
\oauthor{\bsnm{Maceiras}, \binits{J.}},
\oauthor{\bsnm{Abbet}, \binits{P.}},
\oauthor{\bsnm{Racca}, \binits{M.}},
\oauthor{\bsnm{Girgin}, \binits{H.}},
\oauthor{\bsnm{Calinon}, \binits{S.}}:
Online and offline robot programming via augmented reality workspaces.
arXiv preprint arXiv:2107.01884
(2021)
\end{botherref}
\endbibitem

\bibitem{liang2021iropro}
\begin{botherref}
\oauthor{\bsnm{Liang}, \binits{Y.S.}},
\oauthor{\bsnm{Pellier}, \binits{D.}},
\oauthor{\bsnm{Fiorino}, \binits{H.}},
\oauthor{\bsnm{Pesty}, \binits{S.}}:
iropro: An interactive robot programming framework.
International Journal of Social Robotics,
1--15
(2021)
\end{botherref}
\endbibitem

\bibitem{rodamilans2016comparison}
\begin{botherref}
\oauthor{\bsnm{Rodamilans}, \binits{G.B.}},
\oauthor{\bsnm{Villani}, \binits{E.}},
\oauthor{\bsnm{Trabasso}, \binits{L.G.}},
\oauthor{\bparticle{de} \bsnm{Oliveira}, \binits{W.R.}},
\oauthor{\bsnm{Suterio}, \binits{R.}}:
A comparison of industrial robots interface: force guidance system and teach
  pendant operation.
Industrial Robot: An International Journal
(2016)
\end{botherref}
\endbibitem

\bibitem{perzylo2016intuitive}
\begin{bchapter}
\bauthor{\bsnm{Perzylo}, \binits{A.}},
\bauthor{\bsnm{Somani}, \binits{N.}},
\bauthor{\bsnm{Profanter}, \binits{S.}},
\bauthor{\bsnm{Kessler}, \binits{I.}},
\bauthor{\bsnm{Rickert}, \binits{M.}},
\bauthor{\bsnm{Knoll}, \binits{A.}}:
\bctitle{Intuitive instruction of industrial robots: Semantic process
  descriptions for small lot production}.
In: \bbtitle{2016 Ieee/rsj International Conference on Intelligent Robots and
  Systems (iros)},
pp. \bfpage{2293}--\blpage{2300}
(\byear{2016}).
\bcomment{IEEE}
\end{bchapter}
\endbibitem

\bibitem{ladeira2021robmex}
\begin{barticle}
\bauthor{\bsnm{Ladeira}, \binits{M.}},
\bauthor{\bsnm{Ouhammou}, \binits{Y.}},
\bauthor{\bsnm{Grolleau}, \binits{E.}}:
\batitle{Robmex: Ros-based modelling framework for end-users and experts}.
\bjtitle{Journal of Systems Architecture}
\bvolume{117},
\bfpage{102089}
(\byear{2021})
\end{barticle}
\endbibitem

\bibitem{gadre2019end}
\begin{bchapter}
\bauthor{\bsnm{Gadre}, \binits{S.Y.}},
\bauthor{\bsnm{Rosen}, \binits{E.}},
\bauthor{\bsnm{Chien}, \binits{G.}},
\bauthor{\bsnm{Phillips}, \binits{E.}},
\bauthor{\bsnm{Tellex}, \binits{S.}},
\bauthor{\bsnm{Konidaris}, \binits{G.}}:
\bctitle{End-user robot programming using mixed reality}.
In: \bbtitle{2019 International Conference on Robotics and Automation (ICRA)},
pp. \bfpage{2707}--\blpage{2713}
(\byear{2019}).
\bcomment{IEEE}
\end{bchapter}
\endbibitem

\bibitem{materna2018interactive}
\begin{bchapter}
\bauthor{\bsnm{Materna}, \binits{Z.}},
\bauthor{\bsnm{Kapinus}, \binits{M.}},
\bauthor{\bsnm{Beran}, \binits{V.}},
\bauthor{\bsnm{Smr{\v{z}}}, \binits{P.}},
\bauthor{\bsnm{Zem{\v{c}}{\'\i}k}, \binits{P.}}:
\bctitle{Interactive spatial augmented reality in collaborative robot
  programming: User experience evaluation}.
In: \bbtitle{2018 27th IEEE International Symposium on Robot and Human
  Interactive Communication (RO-MAN)},
pp. \bfpage{80}--\blpage{87}
(\byear{2018}).
\bcomment{IEEE}
\end{bchapter}
\endbibitem

\bibitem{bambusek2019combining}
\begin{bchapter}
\bauthor{\bsnm{Bambu{\^s}ek}, \binits{D.}},
\bauthor{\bsnm{Materna}, \binits{Z.}},
\bauthor{\bsnm{Kapinus}, \binits{M.}},
\bauthor{\bsnm{Beran}, \binits{V.}},
\bauthor{\bsnm{Smr{\v{z}}}, \binits{P.}}:
\bctitle{Combining interactive spatial augmented reality with head-mounted
  display for end-user collaborative robot programming}.
In: \bbtitle{2019 28th IEEE International Conference on Robot and Human
  Interactive Communication (RO-MAN)},
pp. \bfpage{1}--\blpage{8}
(\byear{2019}).
\bcomment{IEEE}
\end{bchapter}
\endbibitem

\bibitem{kapinus2020end}
\begin{bchapter}
\bauthor{\bsnm{Kapinus}, \binits{M.}},
\bauthor{\bsnm{Materna}, \binits{Z.}},
\bauthor{\bsnm{Bambu{\v{s}}ek}, \binits{D.}},
\bauthor{\bsnm{Beran}, \binits{V.}}:
\bctitle{End-user robot programming case study: Augmented reality vs. teach
  pendant}.
In: \bbtitle{Companion of the 2020 ACM/IEEE International Conference on
  Human-Robot Interaction},
pp. \bfpage{281}--\blpage{283}
(\byear{2020})
\end{bchapter}
\endbibitem

\bibitem{romero2018speeded}
\begin{barticle}
\bauthor{\bsnm{Romero-Ramirez}, \binits{F.J.}},
\bauthor{\bsnm{Mu{\~n}oz-Salinas}, \binits{R.}},
\bauthor{\bsnm{Medina-Carnicer}, \binits{R.}}:
\batitle{Speeded up detection of squared fiducial markers}.
\bjtitle{Image and vision Computing}
\bvolume{76},
\bfpage{38}--\blpage{47}
(\byear{2018})
\end{barticle}
\endbibitem

\bibitem{markley2007averaging}
\begin{barticle}
\bauthor{\bsnm{Markley}, \binits{F.L.}},
\bauthor{\bsnm{Cheng}, \binits{Y.}},
\bauthor{\bsnm{Crassidis}, \binits{J.L.}},
\bauthor{\bsnm{Oshman}, \binits{Y.}}:
\batitle{Averaging quaternions}.
\bjtitle{Journal of Guidance, Control, and Dynamics}
\bvolume{30}(\bissue{4}),
\bfpage{1193}--\blpage{1197}
(\byear{2007})
\end{barticle}
\endbibitem

\bibitem{kapinus2019spatially}
\begin{bchapter}
\bauthor{\bsnm{Kapinus}, \binits{M.}},
\bauthor{\bsnm{Beran}, \binits{V.}},
\bauthor{\bsnm{Materna}, \binits{Z.}},
\bauthor{\bsnm{Bambu{\v{s}}ek}, \binits{D.}}:
\bctitle{Spatially situated end-user robot programming in augmented reality}.
In: \bbtitle{2019 28th IEEE International Conference on Robot and Human
  Interactive Communication (RO-MAN)},
pp. \bfpage{1}--\blpage{8}
(\byear{2019}).
\bcomment{IEEE}
\end{bchapter}
\endbibitem

\bibitem{brooke1996sus}
\begin{barticle}
\bauthor{\bsnm{Brooke}, \binits{J.}}, \betal:
\batitle{Sus-a quick and dirty usability scale}.
\bjtitle{Usability evaluation in industry}
\bvolume{189}(\bissue{194}),
\bfpage{4}--\blpage{7}
(\byear{1996})
\end{barticle}
\endbibitem

\bibitem{schrep2015ueq}
\begin{botherref}
\oauthor{\bsnm{Schrepp}, \binits{M.}}:
User experience questionnaire handbook
(2015).
\doiurl{10.13140/RG.2.1.2815.0245}
\end{botherref}
\endbibitem

\bibitem{hart1988nasatlx}
\begin{barticle}
\bauthor{\bsnm{Hart}, \binits{S.G.}},
\bauthor{\bsnm{Staveland}, \binits{L.E.}}:
\batitle{Development of nasa-tlx (task load index): Results of empirical and
  theoretical research}.
\bjtitle{Advances in psychology}
\bvolume{52},
\bfpage{139}--\blpage{183}
(\byear{1988})
\end{barticle}
\endbibitem

\bibitem{grier2016}
\begin{barticle}
\bauthor{\bsnm{Grier}, \binits{R.A.}}:
\batitle{How high is high? a meta-analysis of nasa-tlx global workload scores}.
\bjtitle{Proceedings of the Human Factors and Ergonomics Society Annual
  Meeting}
\bvolume{59}(\bissue{1}),
\bfpage{1727}--\blpage{1731}
(\byear{2015})
\end{barticle}
\endbibitem

\bibitem{kapinus2022improved}
\begin{bchapter}
\bauthor{\bsnm{Kapinus}, \binits{M.}},
\bauthor{\bsnm{Bambu{\v{s}}ek}, \binits{D.}},
\bauthor{\bsnm{Materna}, \binits{Z.}},
\bauthor{\bsnm{Beran}, \binits{V.}},
\bauthor{\bsnm{Smr{\v{z}}}, \binits{P.}}:
\bctitle{Improved indirect virtual objects selection methods for cluttered
  augmented reality environments on mobile devices}.
In: \bbtitle{Proceedings of the 2022 ACM/IEEE International Conference on
  Human-Robot Interaction},
pp. \bfpage{834}--\blpage{838}
(\byear{2022})
\end{bchapter}
\endbibitem

\bibitem{bambusek2022handheld}
\begin{bchapter}
\bauthor{\bsnm{Bambu{\v{s}}ek}, \binits{D.}},
\bauthor{\bsnm{Materna}, \binits{Z.}},
\bauthor{\bsnm{Kapinus}, \binits{M.}},
\bauthor{\bsnm{Beran}, \binits{V.}},
\bauthor{\bsnm{Smr{\v{z}}}, \binits{P.}}:
\bctitle{Handheld augmented reality: Overcoming reachability limitations by
  enabling temporal switching to virtual reality}.
In: \bbtitle{Proceedings of the 2022 ACM/IEEE International Conference on
  Human-Robot Interaction},
pp. \bfpage{698}--\blpage{702}
(\byear{2022})
\end{bchapter}
\endbibitem

\bibitem{Tatzgern2016}
\begin{bchapter}
\bauthor{\bsnm{Tatzgern}, \binits{M.}},
\bauthor{\bsnm{Orso}, \binits{V.}},
\bauthor{\bsnm{Kalkofen}, \binits{D.}},
\bauthor{\bsnm{Jacucci}, \binits{G.}},
\bauthor{\bsnm{Gamberini}, \binits{L.}},
\bauthor{\bsnm{Schmalstieg}, \binits{D.}}:
\bctitle{Adaptive information density for augmented reality displays}.
In: \bbtitle{2016 IEEE Virtual Reality (VR)},
pp. \bfpage{83}--\blpage{92}
(\byear{2016}).
\doiurl{10.1109/VR.2016.7504691}
\end{bchapter}
\endbibitem

\bibitem{Veas2008}
\begin{bchapter}
\bauthor{\bsnm{Veas}, \binits{E.}},
\bauthor{\bsnm{Kruijff}, \binits{E.}}:
\bctitle{Vesp’r: design and evaluation of a handheld ar device}.
In: \bbtitle{2008 7th IEEE/ACM International Symposium on Mixed and Augmented
  Reality},
pp. \bfpage{43}--\blpage{52}
(\byear{2008}).
\doiurl{10.1109/ISMAR.2008.4637322}
\end{bchapter}
\endbibitem

\end{thebibliography}


\end{document}